\documentclass[10pt, a4paper]{article}

\usepackage[final]{lrec2026} 

\usepackage{multirow}
\usepackage{tabularx}
\usepackage{booktabs}
\usepackage{graphicx}
\usepackage{subcaption}
\newcommand{\footURL}[1]{\footnote{\url{#1}}}

\title{From Phonemes to Meaning: \\ Evaluating Large Language Models on Tamil}

\name{\begin{tabular}{c c c}
Jeyarajalingam Varsha & Menan Velayuthan & Sumirtha Karunakaran  \\  Rasan Nivethiga & 
\multicolumn{2}{c}{Kengatharaiyer Sarveswaran}
\end{tabular} } 

\address{Department of Computer Science, University of Jaffna, Sri Lanka.
         \\ \{varshajeyaraj, vmenan95, sumirthakarunakaran96,rasanniksha\}@gmail.com, sarves@univ.jfn.ac.lk}

\abstract{
Large Language Models (LLMs) have shown strong generalization across tasks in high-resource languages; however, their linguistic competence in low-resource and morphologically rich languages such as Tamil remains largely unexplored. Existing multilingual benchmarks often rely on translated English datasets, failing to capture the linguistic and cultural nuances of the target language. To address this gap, we introduce \textit{ILAKKANAM}, the first Tamil-specific linguistic evaluation benchmark manually curated using 820 questions from Sri Lankan school-level Tamil subject examination papers. Each question is annotated by trained linguists under five linguistic categories and a factual knowledge category, spanning Grades 1–13 to ensure broad linguistic coverage. We evaluate both closed-source and open-source LLMs using a standardized evaluation framework.  Our results show that Gemini 2.5 achieves the highest overall performance, while open-source models lag behind, highlighting the gap in linguistic grounding. Category- and grade-wise analyses reveal that all models perform well on lower-grade questions but show a clear decline as linguistic complexity increases. Further, no strong correlation is observed between a model’s overall performance and its ability to identify linguistic categories, suggesting that performance may be driven by exposure rather than genuine understanding. 
\\ \newline  }

\begin{document}

\maketitleabstract

\section{Introduction}

Since the public release of ChatGPT in 2022~\citep{OpenAI2024}, Large Language Models (LLMs) have drawn significant public attention and rapidly integrated into everyday life.
This growing interest has attracted substantial investment and funding toward companies developing these systems, resulting in a proliferation of models from different vendors. Closed-source models such as GPT-5~\citep{openaiIntroducingGPT5}, Claude Sonnet~4.5~\citep{anthropicIntroducingClaude}, and Gemini~2.5~\citep{comanici2025gemini25pushingfrontier}, as well as open-source counterparts like LLaMA~4~\citep{metaLlamaHerd}, DeepSeek-V3~\citep{deepseekai2025deepseekv3technicalreport}, Qwen~2.5~\citep{qwen2025qwen25technicalreport}, and Grok~4~\citep{xGrok}, represent this expanding ecosystem. As LLMs become integrated into human workflows, including their use as evaluators for complex tasks (LLM-as-a-Judge)~\citep{gu2025surveyllmasajudge,fu2025reliablemultilingualllmasajudge}, the responsibility lies with the research community to evaluate these models, understand their capabilities, and identify their limitations~\citep{chang2023surveyevaluationlargelanguage}.

The GLUE benchmark~\citep{wang2019gluemultitaskbenchmarkanalysis} and its extended version, SuperGLUE~\citep{wang2019superglue}, established a standardized framework for evaluating language understanding across lexical semantics, logic, and grammar. BLiMP~\citep{warstadt-etal-2020-blimp-benchmark} extended this direction by introducing minimal pair evaluations for core syntactic phenomena such as subject–verb agreement and filler–gap dependencies. HELM~\citep{liang2023holistic} broadened the evaluation scope to incorporate ethical and demographic dimensions while emphasizing the limited multilingual and typological coverage of existing benchmarks. The MMLU dataset~\citep{hendryckstest2021} introduced a multitask evaluation across 57 academic and professional subject domains assessing both factual knowledge and reasoning ability. Despite measurable gains from larger models such as GPT-3, performance remains below expert level, indicating persistent limitations in knowledge depth and reliability.

While several efforts have attempted to create multilingual benchmarks, most are direct translations of their English counterparts~\citep{singh-etal-2025-global,bandarkar-etal-2024-belebele}. Such approaches often fail to capture the cultural and linguistic nuances of the target languages~\citep{Ji_Ji_Bouillon_Seligman_2023}. Following the design of MMLU, comparable benchmarks have been developed for other languages, including Sinhala~\citep{pramodya2025sinhalammlucomprehensivebenchmarkevaluating}, Arabic~\citep{koto-etal-2024-arabicmmlu}, Chinese~\citep{li-etal-2024-cmmlu}, Turkish~\citep{yuksel-etal-2024-turkishmmlu}, Indonesian~\citep{koto-etal-2023-large}, Korean~\citep{son-etal-2025-kmmlu}, and Persian~\citep{Ghahroodi2024KhayyamC}. 
These efforts demonstrate the importance of language-specific benchmarks developed by native-speaking communities, ensuring that linguistic and cultural characteristics are represented, which are otherwise often absent in large-scale multilingual settings.

While SEA-HELM\footnote{Previously known as BHASHA}~\citep{susanto-etal-2025-sea} evaluates the linguistic capabilities of LLMs through its LINDSEA suite for Tamil, its coverage remains limited.
To address this gap, we introduce the \textit{ILAKKANAM}, a manually curated dataset designed for Tamil linguistic assessment. Inspired by MMLU~\citep{hendryckstest2021}, we compile questions from Sri Lankan school-level Tamil language examination papers. 

ILAKKANAM comprises 820 questions spanning Grades 1–13, each annotated by trained linguists under five linguistic categories. This paper outlines the procedures used to collect, clean, annotate, and evaluate these questions against both closed- and open-source LLMs. To summarize, our work makes the following core contributions:

\begin{itemize}
    \item We introduce \textit{ILAKKANAM}, a manually curated Tamil linguistic benchmark consisting of 820 questions from Sri Lankan school-level Tamil language examination papers, annotated across five linguistic categories and a factual knowledge category.  
    \item We design a structured evaluation pipeline to assess both closed-source and open-source LLMs, enabling fine-grained comparison across linguistic dimensions and grade-level complexity.  
    \item Through comprehensive analysis, we reveal a consistent performance gap between closed- and open-source models, limited correlation between linguistic accuracy and category classification, and highlight the need for deeper linguistic grounding in Tamil language modeling. 
\end{itemize}

\section{Background}
This section provides a brief introduction to the Tamil language and the Sri Lankan education system to contextualise and support the work presented in this paper.

\subsection{The Tamil language}
Tamil (tam) is a member of the South Dravidian branch of the Dravidian language family and is spoken by approximately 90 million people worldwide\footURL{https://www.worlddata.info/languages/tamil.php}. It is an agglutinative language with rich morphological and syntactic constructions. Tamil has a documented history of over two millennia, evolving through distinct historical stages. It holds official language status in Sri Lanka, Singapore, and the Indian state of Tamil Nadu, and recognised as a second language in many other countries.

From a computational perspective, Tamil is classified as a low-resource language \cite{abirami-etal-2024-aalamaram} according to Joshi’s typology \cite{joshi-etal-2020-state}. Although widely spoken, Tamil lacks comprehensive, high-quality, and error-free resources—particularly annotated datasets and standardized benchmarks—which limits the development and evaluation of robust NLP tools for the language.

\subsection{The Sri Lankan education system}

Sri Lanka provides 13 years of free general education across four cycles: Primary (Grades 1–5, ages 5–10), Junior Secondary (Grades 6–9, ages 11–14), Senior Secondary (Grades 10–11, ages 15–16), and Advanced Level (Grades 12–13, ages 17–18)\cite{liyanage2014education}. Schooling is compulsory from Grade 1 to 13, with three major national examinations: the Grade 5 Scholarship Exam (for merit-based school access and financial aid), the GCE O/L at Grade 11 (stream selection for A/L), and the GCE A/L at Grade 13 (university admission)\cite{liyanage2014education}.

Tamil is taught both as a subject and a medium of instruction for Tamil-speaking students from Grade 1 to 11, and as a specialization in Grades 12–13. This extended exposure, reinforced by national assessments, builds strong linguistic competence.

By Grade 5\footURL{https://nie.lk/pdffiles/tg/tGR05TG TAMILLANGUAGE.pdf}, students are expected to acquire foundational skills in listening, reading, speaking, and writing. By Grade 11\footURL{https://nie.lk/pdffiles/tg/t11tim159.pdf}, they are expected to master grammar and apply Tamil in academic and social contexts. Though not always explicitly labeled, the curriculum gradually introduces key linguistic domains: phonetics, phonology, morphology, syntax, and semantics. In Grades 12–13, students specializing in Tamil engage with advanced material, including literature, history, and poetry.

\section{Related Work}

\subsection{Tamil in Multilingual Benchmarks}

A few multilingual benchmark evaluation datasets include Tamil among other Indic and Southeast Asian languages.

SEA-HELM\footnote{Previously known as BHASHA}~\citep{susanto-etal-2025-sea,leong2023bhasaholisticsoutheastasian}, through its LINDSEA suite for Tamil, assesses morphology, syntax, and semantics using minimal pair structures and handcrafted diagnostics. In their work, the authors show that both GPT-4 and GPT-3.5-Turbo perform poorly on Tamil morphological analysis, with scores of 16.43\% and 41.43\%, respectively, even when prompted in English. Issues were particularly acute in processing gender, person, and tense agreement, as well as case marking in Question Answering tasks. This highlights that even prominent commercial models still have progress to make in achieving true multilinguality. SEA-HELM further revealed that Tamil’s non-Latin script introduces complications in prompt design, such as the use of capitalization or punctuation conventions that do not align with Tamil’s orthographic norms. 

While PARIKSHA~\citep{watts2024parikshalargescaleinvestigationhumanllm} scored Tamil-generated content highly in terms of fluency and grammaticality, it remains unclear whether this reflects deep language modeling or surface-level token fluency. Collectively, these studies provide strong evidence for the creation and evaluation of LLMs on Tamil in a more fine-grained manner.

\subsection{Categorization of Linguistic Evaluation}

Linguistic evaluation can span several axes. However, we use the five key axes: Phonetics, Phonology, Morphology, Syntax and Semantics for our evaluation\footURL{https://linguistics.ucla.edu/undergraduate/what-is-linguistics/}.


\subsubsection{Phonetics \& Phonology}

Studies outside Tamil, such as \citet{Begus_2025}, evaluated metalinguistic phonological abilities and showed that LLMs generalize over patterns like intervocalic gemination using synthetic words. PolyBench~\citep{suvarna-etal-2024-phonologybench} further introduced tasks on grapheme-to-phoneme conversion and syllable counting, noting that phonological competence remains underexplored even for English. These approaches provide useful reference points for designing Tamil-specific phonological evaluations.

Model evaluations have yet to examine Tamil phonetics and phonology in detail, although some existing studies offer transferable insights. In SEA-HELM’s LINDSEA suite, models such as GPT-4 struggled with verbal reduplication and other syllabic patterns unique to Tamil, producing incoherent translations in modally complex sentences. The same study reported that Tamil’s script posed challenges for grapheme-level prompt formatting, adding further difficulty for phonology-related tasks.

\subsubsection{Morphology}

Tamil’s morphological richness continues to pose difficulties for current LLMs. Results from the LINDSEA tests in SEA-HELM showed frequent errors in morphological agreement, including mismatches in case, gender, and number~\citep{leong2023bhasaholisticsoutheastasian,susanto-etal-2025-sea}. Such problems appeared consistently across Question Answering (QA) and sentence completion tasks, indicating that the models lack a stable inflection mechanism for agglutinative languages.  

PARIKSHA~\citep{watts2024parikshalargescaleinvestigationhumanllm} reported a perfect acceptability score for GPT-4-Turbo on Tamil texts, but this self-assessment measure does not necessarily capture linguistic depth. The gap between surface acceptability and internal representation highlights the need for systematic probing of morphological understanding in Tamil.

\subsubsection{Syntax}

Tamil syntax, with its free word order and nested predicate structures, presents additional challenges for current models. The syntactic diagnostics in LINDSEA evaluate phenomena such as argument structure and filler–gap dependencies, areas where LLMs often underperform~\citep{leong2023bhasaholisticsoutheastasian}. Following the approach of BLiMP~\citep{warstadt-etal-2020-blimp-benchmark}, these tests use minimal pairs to probe specific syntactic contrasts and expose weaknesses in handling long-distance dependencies or embedded clause scrambling.

Benchmarks like PAWS~\citep{zhang-etal-2019-paws} also inform Tamil syntax evaluation by showing that models tend to rely on surface word overlap rather than syntactic structure—a concern particularly relevant for Tamil, where flexible word order is grammatical. \citet{Begus_2025} reported similar limitations, noting that LLMs struggle with recursion and structural ambiguity, consistent with the syntactic difficulties observed for Tamil.

\subsubsection{Semantics}

Semantic understanding remains one of the most challenging aspects for Tamil, both in isolation and in inference-based settings. SEA-HELM evaluated Tamil through the IndicXNLI benchmark and reported that GPT-4o reached a score of 64.7, notably lower than its performance on Vietnamese and Indonesian~\citep{susanto-etal-2025-sea}. The earlier BHASHA study observed even lower semantic comprehension (33.43\%) for GPT-4 when English prompts were used~\citep{leong2023bhasaholisticsoutheastasian}. The model performed better, around 70\%, on salient elements such as the emphatic particle \textit{taan}, suggesting that high-salience, language-specific tokens are more reliably captured.


\section{Data Creation}

\textit{ILAKKANAM} comprises 820 questions spanning Grades 1–13, each annotated by two trained linguists under five linguistic categories (see Table \ref{tab:design-decisions}). Questions targeting factual knowledge are labeled as Facts to evaluate Tamil world knowledge in LLMs. This resource is designed to systematically assess both linguistic competence and culturally grounded knowledge in Tamil.

To build this evaluation resource, we developed a structured pipeline to convert school-level examination materials into a machine-readable dataset. The workflow involves three key stages: (1) collecting question papers from open educational archives, (2) digitizing and cleaning scanned documents, and (3) structuring and filtering the finalized data for experimental use. 

\begin{table*}[!tbp]
\centering
\caption{Design fields and their detailed structure.}
\label{tab:design-decisions}
\renewcommand{\arraystretch}{1.15} 
\resizebox{0.8\textwidth}{!}{    
\begin{tabularx}{\textwidth}{l l X}
\hline
\textbf{Field} & \textbf{Subfield} & \textbf{Description} \\
\hline
Paper ID & — & A composite identifier containing the school or institution name, grade level, and year of examination, enabling precise tracking of question origins. \\
\hline
\multirow{9}{*}{Question Type ID} & QT01 & Fill in the blanks \\
\cline{2-3}
& QT02 & Provide answer based on the given set of letters/words \\
\cline{2-3}
& QT03 & Order the words/letters \\
\cline{2-3}
& QT04 & Question and Answer \\
\cline{2-3}
& QT05 & Sentence completion \\
\cline{2-3}
& QT06 & Rewrite with punctuation marks \\
\cline{2-3}
& QT07 & Multiple Choice Questions (MCQ) \\
\cline{2-3}
& QT08 & Question and Answer based on given paragraph \\
\cline{2-3}
& QT09 & True or False \\
\hline
\multirow{7}{*}{Linguistic Category ID} & L1 & Phonetics \\
\cline{2-3}
& L2 & Phonology \\
\cline{2-3}
& L3 & Morphology \\
\cline{2-3}
& L4 & Syntax \\
\cline{2-3}
& L5 & Semantics \\
\cline{2-3}
& F  & Fact (questions testing knowledge of Tamil cultural, historical, or factual information) \\
\hline
Question Text & — & Question from the examination paper. \\
\hline
Answer & — & The ground truth or correct answer, validated by professional linguists. \\
\hline
Score & — & The point value assigned to each question in the original examination paper, preserved to maintain the weighted importance of different questions. \\
\hline
\end{tabularx}
}
\end{table*}

\subsection{Data Collection}

The dataset was built using school-level Tamil language examination questions from Grades~1–13 in Sri Lanka. The papers were sourced from the \texttt{Noolaham School}\footURL{https://noolaham.school/} section of \texttt{Noolaham.org}\footnote{A digital archive of open Tamil educational resources.}. Two latest exam papers were selected per grade during the curation phase to ensure grade-wise coverage and topic diversity. In the digitization process, essay-type and structured questions were excluded, focusing instead on items that yield concrete, verifiable answers.

\subsection{Digitization \& Cleaning}

The exam papers were available only as scanned PDFs because most were typed using non-Unicode Tamil fonts. To obtain machine-readable text, Optical Character Recognition (OCR) was applied using Google Docs\footURL{https://docs.google.com/document/}, chosen for its accessibility and ease of use. Each file was opened directly in Google Docs, which extracted text through its built-in OCR system.

Automated conversion introduced character, spacing, and formatting errors, which were manually corrected. A simple web interface was used to input cleaned questions and ensure the formatting consistency.

\subsection{Data Structure \& Filtering}

To ensure broader task diversity, the dataset includes nine question types, extending beyond traditional Multiple-Choice Questions (MCQs). After automated filtration and manual correction, a final inspection was performed to remove both exact and near-duplicate questions.

Each datapoint was described using six key fields to ensure comprehensive information capture. A detailed description of these fields and their subfields is provided in Table~\ref{tab:design-decisions}. Questions requiring lengthy written responses were adapted into multiple-choice format with non-obvious answer options to facilitate automated evaluation. 

It should also be noted that questions related to \textit{Pragmatics}, \textit{Discourse}, \textit{Stylistics} and other higher order were incorporated into the \textit{Semantics} category (L5; see Table~\ref{tab:design-decisions}).


Additionally, the assigned marks for each question were incorporated, as higher marks indicate greater complexity. These weighted scores will be useful for model evaluation.

After the extraction and validation phases were completed, the finalized dataset was exported in JSON format for experimental use.

\section{Large Language Model Evaluation}

This section outlines the setup, configuration, and methodology used to evaluate multiple LLMs on the curated Tamil question–answer dataset.

\subsection{Evaluation Setup and Configuration}
We utilized \texttt{Abacus.AI}\footURL{https://chatllm.abacus.ai} as a unified interface to access both open-source and closed-source models. The list of evaluated models can be found in Table~\ref{tab:model-catalog}. 

To guarantee consistency and factuality in responses, the generation temperature was fixed at $0$. All other hyperparameters were left at their default values provided by the platform. To prevent information leakage, ground-truth answers were stored in a separate JSON file, locally, leaving only the questions accessible to the models. Each question file was passed systematically to every LLM under evaluation, and the generated responses were stored in separate files for later analysis.  
This setup allowed multiple models to be evaluated in parallel while preserving data integrity throughout the process.

{\small

\begin{table}[!h]
\centering
\caption{Model catalog grouped by Access and Provider.}
\label{tab:model-catalog}
\renewcommand{\arraystretch}{1.15}
\resizebox{0.8\linewidth}{!}{
\begin{tabularx}{\columnwidth}{l l X}
\hline
\textbf{Access} & \textbf{Provider} & \textbf{Model} \\
\hline
\multirow{4}{*}{Closed-Source} 
  & \multirow{1}{*}{OpenAI} & GPT-5 \\ \cline{3-3}
  
  & \multirow{1}{*}{Anthropic} & Claude Sonnet 4.5 \\ \cline{3-3}
  & \multirow{1}{*}{Google} & Gemini 2.5 \\ \cline{3-3}
  & xAI                     & Grok 4 \\
\hline
\hline
\multirow{3}{*}{Open-Source} 
  & \multirow{1}{*}{Meta}   &  Llama 4 \\ \cline{3-3}
  & \multirow{1}{*}{DeepSeek} & DeepSeek-V3 \\ \cline{3-3}
  & Alibaba                 & Qwen 2.5 72B \\
\hline
\end{tabularx}
}
\end{table}
}

{\small
\begin{table}[ht]
\caption{We report the overall results of each model through Score Percentage (SP). The numbers in parentheses indicate the actual count of correct answers out of 820 questions. Results of the best performing model is made bold.}
\label{tab:overall-res}
\renewcommand{\arraystretch}{1.25} 
\setlength{\tabcolsep}{8pt}        
\centering
\resizebox{0.7\linewidth}{!}{
\begin{tabular}{l r}
\hline
\textbf{Model} & \textbf{SP (/820)} \\ 
\hline
Claude Sonnet 4.5 & 71.09 (579) \\ 
DeepSeek-V3 & 58.04 (491) \\ 
Gemini 2.5 & \textbf{79.55 (659)} \\ 
Llama 4 & 60.67 (501) \\ 
OpenAI GPT5 & 75.94 (633) \\ 
Qwen 2.5 & 37.93 (320) \\ 
xAI Grok 4 & 78.15 (638) \\ 
\hline
\end{tabular}
}
\end{table}
}

\subsection{Evaluation and Analysis}

Model performance was evaluated by comparing each model’s responses with the validated ground-truth answers, in the our local machines. Scores were measured at several levels of detail to capture both overall and category-specific performance. In addition, we conducted a classification task where models were prompted to assign each question to one of six predefined categories (L1–L5 and F) in Zero-short settings. 


Since not all questions from the original examination papers were included, the number of items differed across grades. To allow fair comparison, the grade-level scores were normalized to a 100-point scale before analysis. We used the following equation to obtain the Score Percentage for each analysis.

\begin{equation}
SP = \frac{S_o}{S_t} \times 100
\end{equation}

where \(S_o\) denotes the total score obtained by the model and \(S_t\) denotes the total attainable score.

\begin{figure*}[ht]
\centering
\begin{subfigure}[t]{0.48\linewidth}
    \centering
    \includegraphics[width=\linewidth]{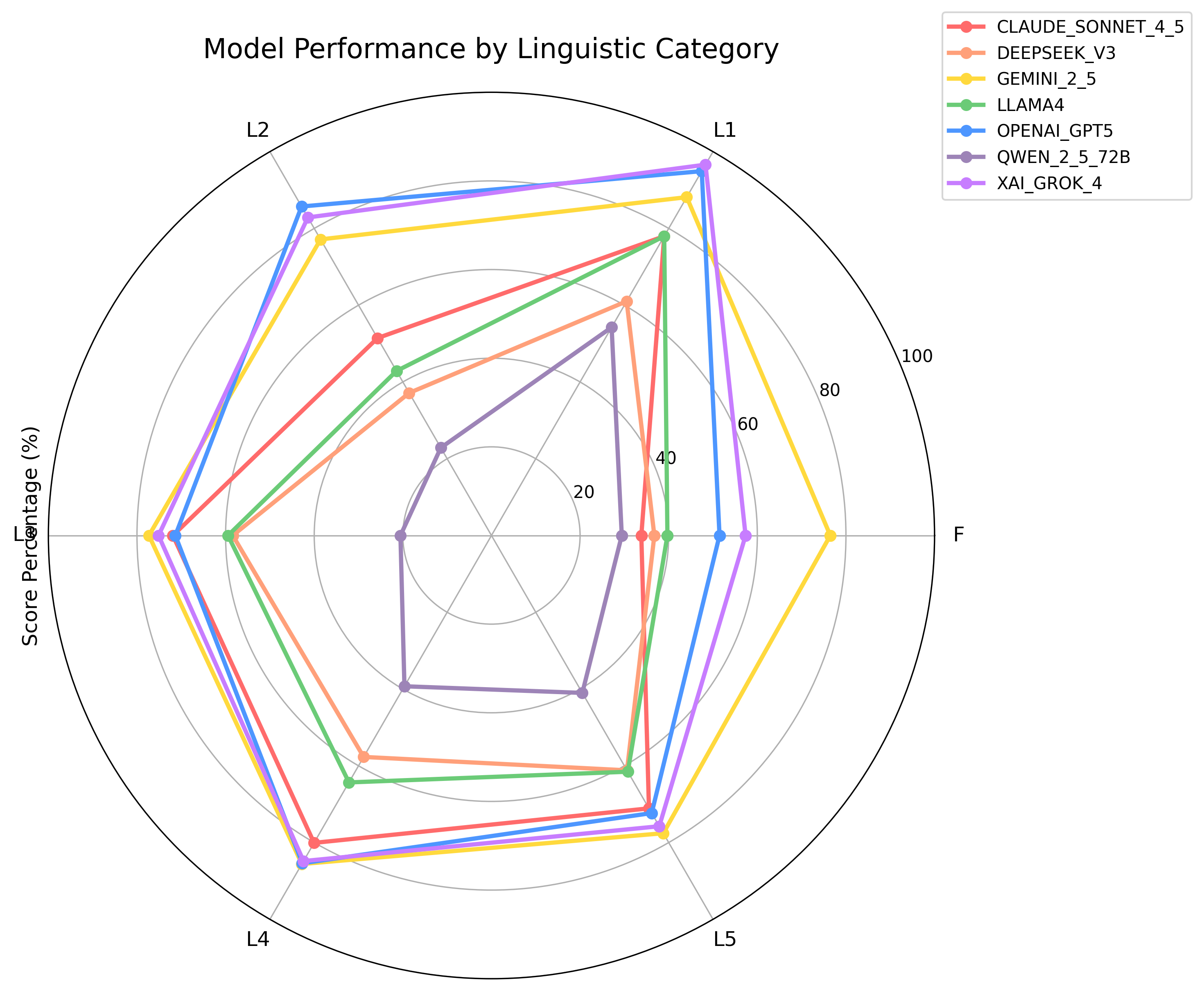}
    \caption{Score Percentage (SP) of each model across linguistic categories.}
    \label{fig:ling-cat}
\end{subfigure}
\hfill
\begin{subfigure}[t]{0.48\linewidth}
    \centering
    \includegraphics[width=\linewidth]{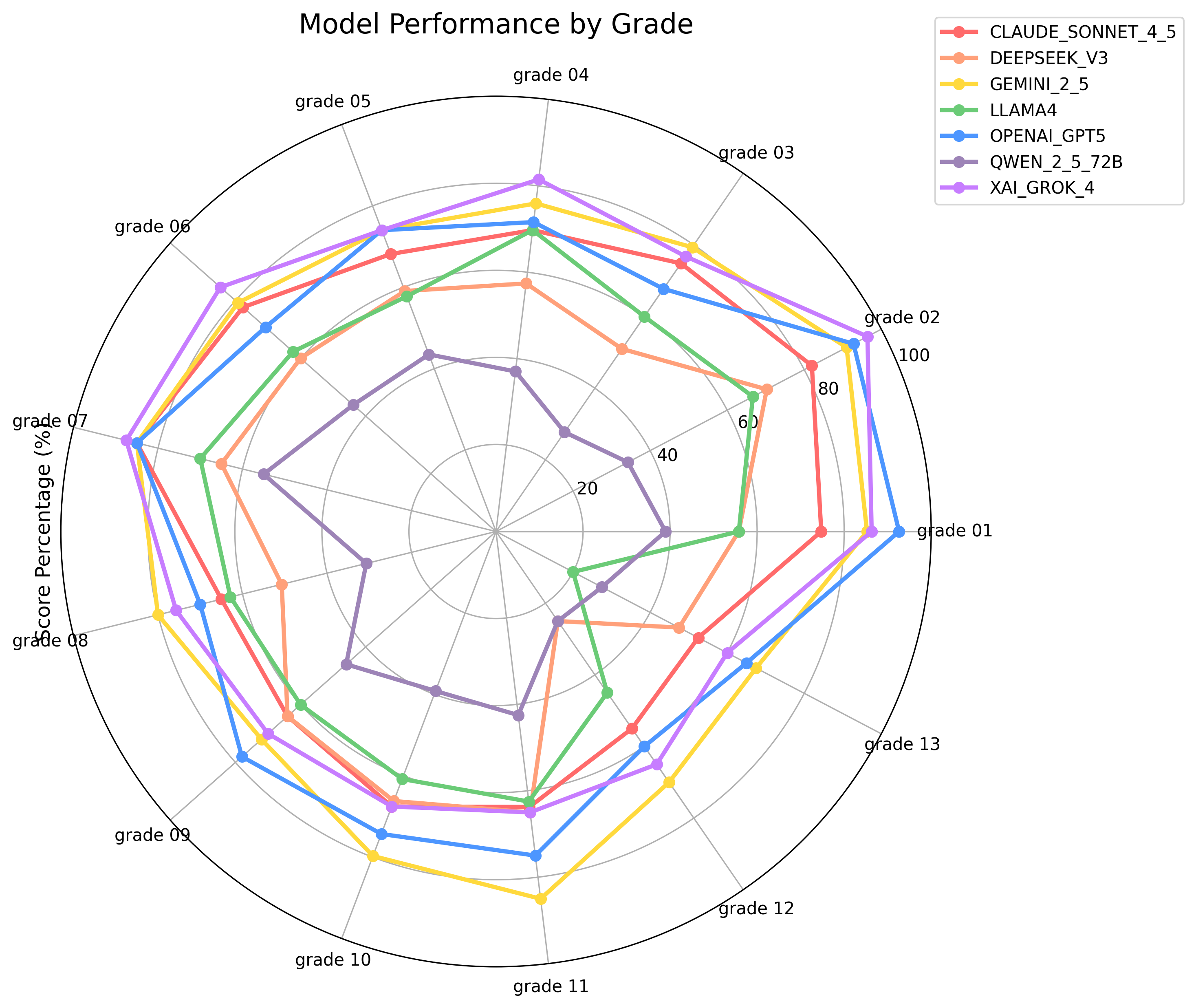}
    \caption{Score Percentage (SP) of each model across grade levels.}
    \label{fig:grade-wise}
\end{subfigure}
\caption{Performance comparison of LLMs. (a) Category-wise results showing linguistic variation. (b) Grade-wise results showing variation across difficulty levels.}
\label{fig:cat-wise}
\end{figure*}

\subsection{Manual Evaluation and Validation}

Responses marked as incorrect in the automated evaluation were separated and manually reviewed by trained linguists. The review focused on two main goals: identifying cases where model outputs differed from the reference but were still linguistically acceptable, and capturing valid alternative answers that were not present in the original ground truth. All such verified alternative responses were subsequently incorporated into the final evaluation metrics, ensuring that the reported accuracy more accurately reflected true model performance rather than strict lexical matching.

\section{Results and Discussion}

\begin{table*}[ht]
\caption{Linguistic category-wise performance of each model through Score Percentage (SP). The numbers in parentheses indicate the actual count of correct answers. The best score in each linguistic category is made bold.}
\label{tab:linguistic-category}
\renewcommand{\arraystretch}{1.25}
\setlength{\tabcolsep}{6pt}
\centering
\resizebox{0.9\linewidth}{!}{
\begin{tabular}{lcccccc}
\hline
\textbf{Model} & \textbf{L1 (20)} & \textbf{L2 (32)} & \textbf{L3 (75)} & \textbf{L4 (169)} & \textbf{L5 (512)} & \textbf{F (12)} \\
\hline
Claude Sonnet 4.5 & 77.97 (15) & 44.74 (15) & 75.34 (53) & 79.28 (129) & 69.68 (362) & 37.50 (5) \\
DeepSeek-V3 & 61.02 (13) & 36.84 (13) & 59.36 (42) & 57.55 (99) & 59.55 (319) & 37.50 (5) \\
Gemini 2.5 & 88.14 (17) & 71.05 (23) & \textbf{79.91 (57)} & \textbf{85.51 (143)} & \textbf{77.69 (410)} & \textbf{75.00 (9)} \\
Llama 4 & 77.97 (15) & 28.95 (9) & 62.10 (45) & 63.78 (105) & 61.12 (323) & 32.50 (4) \\
OpenAI GPT5 & 94.92 (19) & 78.95 (25) & 75.80 (54) & 84.51 (142) & 72.62 (386) & 57.50 (7) \\
Qwen 2.5 (72B) & 54.24 (10) & 31.58 (10) & 21.92 (17) & 40.44 (69) & 39.22 (210) & 35.00 (4) \\
xAI Grok 4 & \textbf{96.61 (19)} & \textbf{84.21 (27)} & 79.00 (56) & 83.90 (136) & 75.56 (393) & 57.50 (7) \\
\hline
Average score & 78.70 & 53.76 & 64.78 & 70.71 & 65.06 & 47.50 \\
\hline
\end{tabular}
}
\end{table*}
We evaluate model performance across four complementary dimensions to obtain a comprehensive understanding of their linguistic and task-level behavior: 
\begin{enumerate}
    \item examine overall performance across the full dataset
    \item analyze results by linguistic category (L1–L5 and F) to capture category-specific variations (refer Table~\ref{tab:linguistic-category} and Figure~\ref{fig:ling-cat})
    \item report grade-wise performance to observe how models handle questions of varying complexity (refer Table~\ref{tab:gradewise-res} and Figure~\ref{fig:grade-wise})
    \item present results from the linguistic category classification task, which assesses the models’ ability to identify the underlying linguistic phenomenon in each question.
\end{enumerate}

\subsection{Overall Performance}
As presented in Table~\ref{tab:overall-res}, the overall evaluation reveals clear variation in performance across models, with Gemini 2.5 achieving the highest score percentage of approximately 80\%, reflecting superior linguistic understanding and factual precision—likely supported by Google’s extensive multilingual and high-quality training data. Among all models, the closed-source group consistently occupies the top three ranks, highlighting their advantage in optimization, alignment, and dataset diversity. Among the open-source models, LLaMA 4 performed comparatively well with a score percentage of 60.67\%, demonstrating strong generalization ability despite limited access to proprietary data. In contrast, Qwen 2.5, despite being a large-scale 72B model, recorded the lowest score (39.02\%), reinforcing that model size alone does not guarantee better performance without effective linguistic grounding and diverse, representative training corpora.
\begin{table*}[ht]
\caption{We report the Score Percentage (SP) for each grade, with the number of correct answers shown in parentheses. The best result in each grade is highlighted in bold.}
\label{tab:gradewise-res}
\renewcommand{\arraystretch}{1.25}
\setlength{\tabcolsep}{6pt}
\centering
\resizebox{\linewidth}{!}{
\begin{tabular}{|c|l|l|l|l|l|l|l|}
\hline
\textbf{Grade} &
\multicolumn{1}{c|}{\rotatebox{90}{\textbf{Claude Sonnet 4.5}}} &
\multicolumn{1}{c|}{\rotatebox{90}{\textbf{DeepSeek-V3}}} &
\multicolumn{1}{c|}{\rotatebox{90}{\textbf{Gemini 2.5}}} &
\multicolumn{1}{c|}{\rotatebox{90}{\textbf{Llama 4}}} &
\multicolumn{1}{c|}{\rotatebox{90}{\textbf{OpenAI GPT5}}} &
\multicolumn{1}{c|}{\rotatebox{90}{\textbf{Qwen 2.5}}} &
\multicolumn{1}{c|}{\rotatebox{90}{\textbf{xAI Grok 4}}} \\
\hline
\textbf{Gr1 (60)}  & 74.74 (41) & 55.79 (29) & 85.26 (50) & 55.79 (29) & \textbf{92.63 (55)} & 38.95 (21) & 86.32 (51) \\
\textbf{Gr2 (58)}  & 81.98 (46) & 70.27 (41) & 90.99 (53) & 66.67 (39) & 92.79 (54) & 34.23 (20) & \textbf{96.40 (56)} \\
\textbf{Gr3 (90)}  & 74.84 (68) & 50.97 (46) & \textbf{79.35 (70)} & 60.00 (55) & 67.74 (62) & 27.74 (26) & 76.77 (71) \\
\textbf{Gr4 (120)}  & 69.75 (91) & 57.41 (76) & 75.93 (98) & 69.75 (89) & 71.60 (94) & 37.04 (48) & \textbf{81.48 (102)} \\
\textbf{Gr5 (111)}  & 68.18 (79) & 59.09 (73) & \textbf{74.03 (89)} & 57.79 (71) & \textbf{74.03 (88)} & 43.51 (52) & \textbf{74.03 (88)} \\
\textbf{Gr6 (37)}  & 77.69 (29) & 60.00 (22) & 79.23 (29) & 62.31 (23) & 70.77 (26) & 43.85 (16) & \textbf{84.62 (31)} \\
\textbf{Gr7 (40)}  & 85.00 (34) & 65.00 (26) & 85.00 (34) & 70.00 (28) & 85.00 (34) & 55.00 (22) & \textbf{87.50 (35)} \\
\textbf{Gr8 (40)}  & 65.00 (26) & 50.71 (21) & \textbf{80.00 (32)} & 62.86 (25) & 70.00 (28) & 30.71 (13) & 75.71 (30) \\
\textbf{Gr9 (50)}  & 64.00 (32) & 64.00 (32) & 72.00 (36) & 60.00 (30) & \textbf{78.00 (39)} & 46.00 (23) & 70.00 (35) \\
\textbf{Gr10 (74)} & 67.57 (50) & 66.22 (49) & \textbf{79.73 (59)} & 60.81 (45) & 74.32 (55) & 39.19 (29) & 67.57 (50) \\
\textbf{Gr11 (80)} & 63.75 (51) & 65.00 (52) & \textbf{85.00 (68)} & 62.50 (50) & 75.00 (60) & 42.50 (34) & 65.00 (52) \\
\textbf{Gr12 (20)} & 55.00 (11) & 25.00 (5) & \textbf{70.00 (14)} & 45.00 (9) & 60.00 (12) & 25.00 (5) & 65.00 (13) \\
\textbf{Gr13 (40)} & 52.50 (21) & 47.50 (19) & \textbf{67.50 (27)} & 20.00 (8) & 65.00 (26) & 27.50 (11) & 60.00 (24) \\
\hline
\end{tabular}
}
\end{table*}

\subsection{Linguistic-wise Evaluation }
Models were also evaluated across linguistic categories to assess how effectively they capture different aspects of linguistic understanding. The results are presented in Table~\ref{tab:linguistic-category} and Figure~\ref{fig:cat-wise}. The best-performing model overall, Gemini 2.5, demonstrated consistent performance across all categories, indicating a balanced grasp of Tamil language structure. In contrast, most other models showed weaker performance in the factual (F) category, which does not fall under linguistic analysis but assesses a model’s Tamil world knowledge—including familiarity with poem authors, cultural references, literary works, and historical facts. When considering the linguistic dimensions alone, none of the models exceeded 80\%, with Gemini 2.5 achieving the highest scores of 79.91\% in L3 and 77.69\% in L5, showing its relative strength in linguistic comprehension. The highest score within a linguistic category was observed in L1 (phonetics)(see Figure~\ref{fig:ling-cat}), achieved by Grok 4 (96.61\%), closely followed by GPT-5 (94.92\%), reflecting their superior performance in this aspect.

In addition, as noted in previous studies, models continue to perform poorly on phonology and morphology tasks, likely due to the complex and rich morphological structure of Tamil. Although better performance is generally expected on semantic tasks—since models can leverage contextual information to infer meaning—the overall scores remain low. This may also be attributed to the fact that the semantic test set also contains pragmatic questions, which introduce additional challenges for Tamil.

\subsection{Grade-wise Evaluation}
Table~\ref{tab:gradewise-res} presents the grade-wise evaluation results of all LLMs, while Figure~\ref{fig:grade-wise} illustrates the overall performance trend across grades. The grade-wise score percentage analysis shows that all models performed relatively well in Grades 1 and 2, which is expected since the questions at these levels are simpler and focus more on basic language skills that can be easily captured from training data. As the grade level increases, particularly from Grade 5 onwards, a noticeable decline in performance is observed across models. This corresponds to the increasing linguistic and conceptual complexity of questions that require a stronger command of Tamil grammar, vocabulary, and linguistic structure. The lowest scores appear around Grades 5 and 13, which align with national-level examinations in Sri Lanka, where the questions are more challenging and require precise linguistic understanding. Even though Grade 11 (G.C.E. O/L) is also a national exam, it is less competitive, which is reflected in slightly better scores. We see that xAI Grok 4 performing best for lower grades 1-7, while Gemini 2.5 outperforms all he models at higher grades (8-13). Among all models, Gemini 2.5 maintained consistently high performance across grades, achieving above 67\% even at the higher levels, reflecting its stronger adaptability to linguistic variation and question complexity compared to other models.
\begin{table}[ht]
\caption{Overall Accuracy Percentage (AP) for the linguistic classification task. Numbers in parentheses show correctly classified tags out of 820. Best-performing model is in bold.}
\label{tab:ling-tag}
\renewcommand{\arraystretch}{1.25} 
\setlength{\tabcolsep}{8pt}        
\centering
\resizebox{0.8\linewidth}{!}{
\begin{tabular}{l r}
\hline
\textbf{Model} & \textbf{SP (/820)} \\
\hline
Claude Sonnet 4.5 & 44.02 (361) \\
DeepSeek V3 & 27.93 (229) \\
Gemini 2.5 & 52.07 (427) \\
LLaMA 4 & 51.59 (423) \\
\textbf{OpenAI GPT-5} & \textbf{65.61 (538)} \\
Qwen 2.5 72B & 52.07 (427) \\
xAI Grok 4 & 61.59 (505)\\
\hline
\end{tabular}
}
\end{table}
\subsection{Linguistic Category Classification}

In addition to the primary evaluation task, an additional experiment was conducted to better understand each model’s ability to capture linguistic awareness (refer Table~\ref{tab:ling-tag}). In this setup, the models were instructed to assign an appropriate linguistic tag (from L1 to L5) to each question, with any item not fitting a linguistic category explicitly directed to be tagged as F. A notable observation emerged from this experiment: while Gemini 2.5 performed exceptionally well in answering questions during the main evaluation, it ranked second in this linguistic categorization task, falling behind GPT-5, with 111 fewer correctly tagged questions. This finding reinforces the argument that Gemini 2.5’s strong performance may not stem from genuine linguistic understanding, but rather from its extensive training data coverage. Since the evaluation questions are based on Tamil school examination papers, they are likely to follow predictable patterns and exhibit limited novelty, making them easier for Gemini to match with seen data. In contrast, the linguistic tagging task requires deeper analytical ability and true understanding of linguistic structures—skills that cannot be derived from memorized or surface-level data. In this respect, GPT-5 demonstrated stronger linguistic awareness and interpretive precision.

\section{Copyrights}

All the questions used in this work have been sourced from publicly available materials that are licensed under the Creative Commons Attribution-NonCommercial-ShareAlike 4.0 International (CC BY-NC-SA 4.0).

\section{Conclusion}

We introduce \textit{ILAKKANAM}, a Tamil linguistic benchmark dataset consisting of 820 manually curated questions from Sri Lankan school-level Tamil grade-wise examination papers. We perform an extensive evaluation on both closed-source and open-source LLMs and show that there is a clear gap between them in terms of Tamil linguistic performance. We analyze the overall results, discuss model performance across linguistic categories and grade levels, and present observations from the linguistic categorization task. Our studies show that Gemini 2.5 performs best on our benchmark dataset. We find no clear relationship between a model’s performance on the linguistic tasks and it's ability to classify these questions into their respective linguistic categories, further suggesting that LLM performance may not stem from linguistic understanding but rather from broad exposure to training data. We hope that both the \textit{ILAKKANAM} dataset and our analyses help researchers better understand the limitations of current models and encourage further efforts toward evaluating and benchmarking LLMs for the Tamil language. Dataset will be provided upon request to ensure no data leakage.

\section{Limitation}
First, only two recent papers were selected from each grade level, which may not fully capture the breadth and diversity of Tamil linguistic phenomena. As a result, certain aspects of language use and structure may not be adequately represented in the current dataset. We are actively expanding the question bank to improve linguistic coverage and representation.

Second, our analysis focused on five core linguistic dimensions: phonetics, phonology, morphology, syntax, and semantics. Extended linguistic areas, such as pragmatics and stylistics, were grouped under semantics because of their limited presence in the dataset. This grouping may reduce the granularity of analysis for these higher-level aspects.

\section{Bibliographical References}\label{sec:reference}

\bibliographystyle{lrec2026-natbib}
\bibliography{references}


\end{document}